\documentclass[11pt]{article}
\usepackage{mtsummit2019}
\usepackage{times}
\usepackage{url}
\usepackage{latexsym}
\usepackage[small,bf]{caption} 
\usepackage{graphicx}
\graphicspath{ {images/} }
\usepackage[caption=false]{subfig}
\usepackage{mathtools}

\usepackage{multirow}
\usepackage{xcolor}

\title{An Intrinsic Nearest Neighbor Analysis\\ of Neural Machine Translation Architectures}

\author{Hamidreza Ghader\\
  Informatics Institute,\\ 
  University of Amsterdam,\\
  The Netherlands\\
  {\tt h.ghader@uva.nl}  \And
  Christof Monz\\
  Informatics Institute,\\ 
  University of Amsterdam,\\
  The Netherlands\\
  {\tt c.monz@uva.nl}}

\date{}

\begin{document}
\maketitle
\begin{abstract}
Earlier approaches indirectly studied the information captured by the hidden states of recurrent and non-recurrent neural machine translation models by feeding them into different classifiers. 
In this paper, we look at the encoder hidden states of both transformer and recurrent machine translation models from the nearest neighbors perspective. We investigate to what extent the nearest neighbors share information with the underlying word embeddings as well as related WordNet entries. Additionally, we study the underlying syntactic structure of the nearest neighbors to shed light on the role of syntactic similarities in bringing the neighbors together. We compare transformer and recurrent models in a more intrinsic way in terms of capturing lexical semantics and syntactic structures, in contrast to extrinsic approaches used by previous works. 
In agreement with the extrinsic evaluations in the earlier works, our experimental results show that transformers are superior in capturing lexical semantics, but not necessarily better in capturing the underlying syntax. Additionally, we show that the backward recurrent layer in a recurrent model learns more about the semantics of words, whereas the forward recurrent layer encodes more context.
\end{abstract}

\section{Introduction}

Neural machine translation (NMT) has achieved state-of-the-art performance for many language pairs \cite{bahdanau-EtAl:2015:ICLR,luong-EtAl:2015:ACL-IJCNLP,jean-EtAl:2015:ACL-IJCNLP,wu2016google,NIPS2017_7181}. Additionally, it is straightforward to train an NMT system in an end-to-end fashion. This has been made possible with an encoder-decoder architecture that encodes the source sentence into a distributed representation and then decodes this representation into a sentence in the target language. While earlier work has investigated what information is captured by the attention mechanism of an NMT system \cite{ghader2017does}, it is not exactly clear what linguistic information from the source sentence is encoded in the hidden distributed representation themselves. Recently, some attempts have been made to shed some light on the information that is being encoded in the intermediate distributed representations \cite{shi-padhi-knight:2016:EMNLP2016,belinkov2017neural}. 

Feeding the hidden states of the encoder of different seq2seq systems, including multiple NMT systems, as the input to different classifiers, \newcite{shi-padhi-knight:2016:EMNLP2016} aim to show what syntactic information is encoded in the hidden states. They provide evidence that syntactic information such as the voice and tense of a sentence and the part-of-speech (POS) tags of words are being learned with reasonable accuracy. They also provide evidence that more complex syntactic information such as the parse tree of a sentence is also learned, but with lower accuracy. 

\newcite{belinkov2017neural} follow the same approach as \newcite{shi-padhi-knight:2016:EMNLP2016} to conduct more analyses about how syntactic and morphological information are encoded in the hidden states of the encoder. They carry out experiments for POS tagging and morphological tagging. They study the effect of different word representations, different layers of the encoder and target languages on the accuracy of their classifiers to reveal the impact of these variables on the amount of the syntactic information captured in the hidden states.

Additionally, there are recent approaches that compare different state-of-the-art encoder-decoder architectures in terms of their capabilities to capture syntactic structures \cite{D18_Recurrent} and lexical semantics \cite{D18-1458_Self-Attention}. These works also use some extrinsic tasks to do the comparison. \newcite{D18_Recurrent} use subject-verb agreement and logical inference tasks to compare recurrent models with transformers. On the other hand, \newcite{D18-1458_Self-Attention} use subject-verb agreement and word sense disambiguation for comparing those architectures in terms of capturing syntax and lexical semantics respectively. In addition to these tasks, \newcite{C18-1054} compare recurrent models with transformers on a multilingual machine translation task.

Despite the approaches discussed above, attempts to study the hidden states more intrinsically are still missing. For example, to the best of our knowledge, there is no work that studies the encoder hidden states from a nearest neighbor perspective to compare these distributed word representations with the underlying word embeddings. It seems intuitive to assume that the hidden state of the encoder corresponding to an input word conveys more contextual information compared to the embedding of the input word itself. But what type of information is captured and how does it differ from the word embeddings? Furthermore, how different is the information captured by different architectures, especially recurrent vs self-attention architectures which use entirely different approaches to capture context?


In this paper, we choose to investigate the hidden states from a nearest neighbors perspective and try to show the similarities and differences between the hidden states and the word embeddings. We collect statistics showing how much information from embeddings of the input words is preserved by the corresponding hidden states. We also try to shed some light on the information encoded in the hidden states that goes beyond what is transferred from the word embeddings. To this end, we analyze how much the nearest neighbors of words based on their hidden state representations are covered by direct relations in WordNet \cite{wordnet_book,Miller:1995:WLD:219717.219748}. For our German side experiments, we use GermaNet~\cite{W97-0802,HENRICH10.264}. From now on, we use \textit{WordNet} to refer to either WordNet or GermaNet.

This paper does not directly seek improvements to neural translation models, but to further our understanding of the inside behaviour of these models. It explains what information is learned in addition to what is already captured by embeddings. This paper makes the following contributions:

\begin{enumerate}
\item We provide interpretable representations of hidden states in NMT systems highlighting the differences between hidden state representations and word embeddings.
\item We compare transformer and recurrent models in a more intrinsic way in terms of capturing lexical semantics and syntactic structures.
\item We provide analyses of the behaviour of the hidden states for each direction layer and the concatenation of the states from the direction layers.
\end{enumerate}

\begin{table*}[thb]
\centering
\small
\begin{tabular}{ccccc}
 \multicolumn{5}{c}{English-German} \\
\hline
 Model&  test2014 & test2015 & test2016 & test2017 \\
\hline
Recurrent & 24.65 & 26.75 & 30.53 & 25.51\\
Transformer & 26.93 & 29.01& 32.44 & 27.36 \\
\hline
\\
 \multicolumn{5}{c}{German-English} \\
\hline
 Model&  test2014 & test2015 & test2016 & test2017 \\
 \hline
 Recurrent & 28.40 & 29.61 & 34.28 & 29.64 \\
Transformer & 30.15 & 30.92 & 35.99 & 31.80 \\
\hline
\end{tabular}
\caption{ Performance of our experimental systems in BLEU on WMT \cite{bojar2017findings} German-English and English-German standard test sets. }
\label{table:Bleu}
\end{table*}

\section{Datasets and Models}

We conduct our analysis using recurrent and transformer machine translation models. Our recurrent model is a two-layer bidirectional recurrent model with Long Short-Term Memory (LSTM) units \cite{Hochreiter:1997} and global attention \cite{DBLP_journals_corr_LuongPM15}. The encoder consists of a two-layer unidirectional forward and a two-layer unidirectional backward pass. The corresponding output representations from each direction are concatenated to form the encoder hidden state representation for each word. A concatenation and down-projection of the last states of the encoder is used to initialize the first state of the decoder. The decoder uses a two-layer unidirectional (forward) LSTM. We use no residual connection in our recurrent model as they have been shown to result in performance drop if used on the encoder side of recurrent model \cite{D17-1151}. Our transformer model is a 6-layer transformer with multi-headed attention of 8 heads \cite{NIPS2017_7181}. We choose these settings to obtain competitive models with the relevant core components from each architecture.

We train our models for two directions, namely English-German and German-English, both of which use the WMT15 parallel training data. We exclude 100k randomly chosen sentence pairs which are used as our held-out data. Our recurrent system has hidden state dimensions of the size of 1,024 (512 for each direction) and is trained using a batch size of 64 sentences. The learning rate is set to 0.001 for the Adam optimizer \cite{kingma:adam} with a maximum gradient norm of 5. A dropout rate of 0.3 has been used to avoid overfitting.
Our transformer model has hidden state dimensions of 512 and a batch size of 4096 tokens and uses layer normalization~\cite{NIPS2017_7181}. A learning rate of 2 changed under warm-up strategy with 8000 warm-up steps is used for Adam optimizer with $\beta_1=0.9$, $\beta_2=0.998$ and $\epsilon=10^{-9}$ \cite{NIPS2017_7181}. The dropout rate is set to 0.1, and no gradient clipping is used. The word embedding size of both models is 512. We apply Byte-Pair Encoding (BPE) \cite{sennrichP16-1162} with 32K merge operation. 

We train our models until convergence and then use the trained models to translate 100K sentences from a held-out dataset and log the hidden states for later use in our analyses. The 100K held-out data is randomly chosen from the WMT15 parallel training data. The remaining of the WMT15 parallel training data is used as our training data.

Table~\ref{table:Bleu} summarizes the performance of our experimental models in BLEU \cite{papineni2002bleu} on different standard test sets. This is to make sure that the models are trustable.

\begin{figure}[thb]
\centering
\includegraphics[scale=0.73]{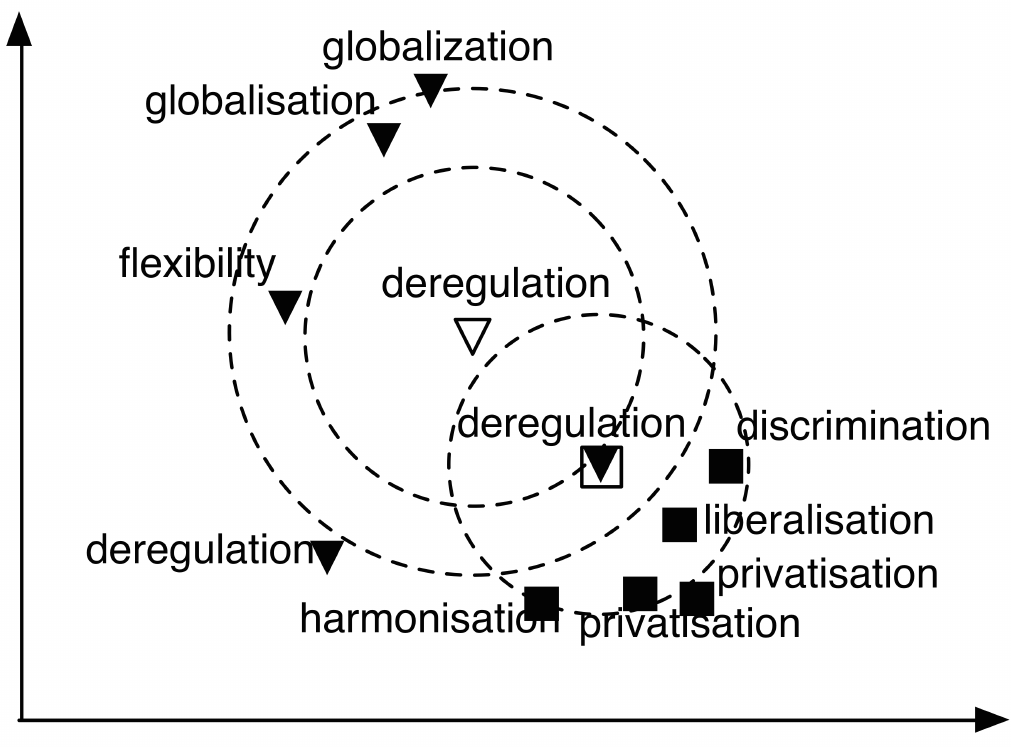}
\caption{An example of 5 nearest neighbors of two different occurrences of the word ``deregulation". Triangles are the nearest neighbors of ``deregulation" shown with the empty triangle. Squares are the nearest neighbors of ``deregulation" shown with the empty square. }
\label{fig:Nearest_Neighbors}
\end{figure}

\section{Nearest Neighbors Analysis}

Following earlier work on word embeddings \cite{mikolov2013distributed,W16-1620}, we choose to look into the nearest neighbors of the hidden state representations to learn more about the information encoded in them.
We treat each hidden state as the representation of the corresponding input token. This way, each occurrence of a word has its own representation. Based on this representation, we compute the list of $n$ nearest neighbors of each word occurrence. We set $n$ equal to 10 in our experiments. Cosine similarity is used as the distance measure.

In the case of our recurrent neural model, we use the concatenation of the corresponding output representations of our two-layer forward and two-layer backward passes as the hidden states of interest for our main experiments. We also use the output representations of the forward and the backward passes for our direction-wise experiments. In the case of our transformer model, we use the corresponding output of the top layer of the encoder for each word as the hidden state representation of the word.

Figure~\ref{fig:Nearest_Neighbors} shows an example of 5 nearest neighbors for two different occurrences of the word ``deregulation". Each item in this figure is a specific word occurrence, but we have removed occurrence information for the sake of simplicity.

\subsection{Hidden States vs Embeddings}
\label{sec:hiddenStatesVsEmbeddings}

Here, we count how many of the words in the nearest neighbors lists of hidden states are covered by the nearest neighbors list based on the corresponding word embeddings. Just like the hidden states, the word embeddings used for computing the nearest neighbors are also from the same system and the same trained model for each experiment. The nearest neighbors of the word embeddings are also computed using cosine similarity. It should be noted that we generate the nearest neighbors lists for the embeddings and the hidden states separately and never compute cosine similarity  between word embeddings and the hidden state representations.

Coverage is formally computed as follows:
\begin{equation}
\label{eq:embeddingCov}
cp^{H,E}_{w_{i,j}} = \frac{\left\vert{C^{H,E}_{w_{i,j}}} \right\vert}{\left\vert{N^{H}_{w_{i,j}}}\right\vert}
\end{equation}
where 
\begin{equation}
\label{eq:hiddenEmbeddingIntersection}
C^{H,E}_{w_{i,j}} = N^{H}_{w_{i,j}} \cap N^{E}_{w}
\end{equation}
and $N^{H}_{w_{i,j}}$ is the set of the $n$ nearest neighbors of word $w$ based on hidden state representations. Since there is a different hidden state for each occurrence of a word, we use $i$ as the index of the sentence of occurrence and $j$ as the index of the word in the sentence. Similarly, $N^{E}_{w}$ is the set of the $n$ nearest neighbors of word $w$, but based on the embeddings. 

Word embeddings tend to capture the dominant sense of a word, even in the presence of significant support for other senses in the training corpus \cite{W16-1620}. Additionally, it is reasonable to assume that a hidden state corresponding to a word occurrence captures more of the current sense of the word. Comparing the lists can provide useful insights as to which hidden state-based neighbours are not strongly related to the corresponding word embedding.
Furthermore, it shows in what cases the dominant information encoded in the hidden states comes from the corresponding word embedding and to what extent other information has been encoded in the hidden state.

\subsection{WordNet Coverage}
In addition, we also compute the coverage of the list of the nearest neighbors of hidden states with the directly related words from WordNet. This can shed further light on the capability of hidden states in terms of learning the sense of the word in the current context. Additionally, it could play the role of an intrinsic measure to compare different architectures in their ability to learn lexical semantics. 
To this end, we check how many words from the nearest neighbors list of a word, based on hidden states, are in the list of related words of the word in WordNet.
More formally, we define $R_w$ to be the union of the sets of synonyms, antonyms, hyponyms and hypernyms of word $w$ in WordNet:

\begin{equation}
\label{eq:wordnetCov}
cp^{H,W}_{w_{i,j}} = \frac{\left\vert{C^{H,W}_{w_{i,j}}}\right\vert}{\left\vert{N^{H}_{w_{i,j}}}\right\vert}
\end{equation}
where
\begin{equation}
\label{eq:hiddenWordnetIntersection}
C^{H,W}_{w_{i,j}} = N^{H}_{w_{i,j}} \cap R_w
\end{equation}
and $N^{H}_{w_{i,j}}$ is the set of the $n$ nearest neighbors of word $w$ based on hidden state representations. 





\begin{figure}[thb]
\centering
\subfloat[$v_{0}$\label{fig:word_of_interest}]{%
\includegraphics[scale=1]{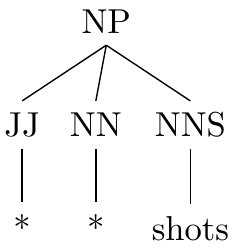}
} \hfill
\subfloat[$v_{1}$ \label{fig:neighbor1}]{%
\includegraphics[scale=1]{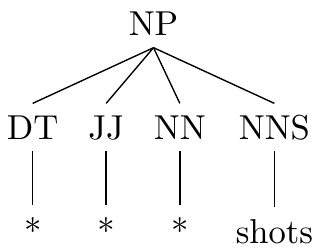}
}

\subfloat[$v_{2}$ \label{fig:neighbor2}]{%
\includegraphics[scale=1]{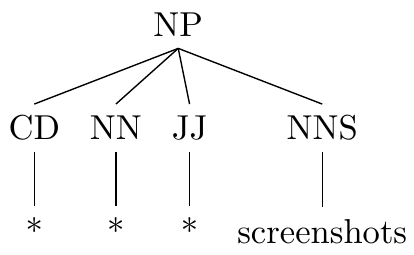}
}\hfill
\subfloat[$v_{10}$\label{fig:neighbor3}]{%
\includegraphics[scale=1]{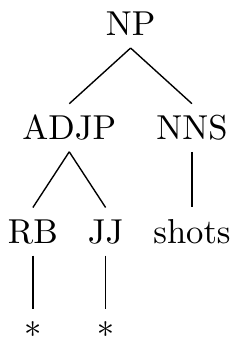}
}
\caption{The figure shows the corresponding word and constituent subtree of query hidden state ($v_{0}$) and the corresponding word and subtree of the first ($v_{1}$) and the second ($v_{2}$) and the last ($v_{10}$) nearest neighbors of it.}
\label{fig:trees}
\end{figure}

\begin{table*}[thb]
\centering
\small
\begin{tabular}{cccccc}
\hline
Model &POS & English-German& $\sigma^2$ & German-English & $\sigma^2$\\
\hline
\multirow{5}{*}{Recurrent} & All POS & 18\%& 4 & 24\% & 7\\
& VERB & 29\% & 5 & 31\%& 5\\
& NOUN & 14\%& 3& 19\%& 8\\
& ADJ & 19\% & 3 & 31\%& 7\\
& ADV & 36\%& 5 &48\%& 2 \\
\hline
\multirow{5}{*}{Transformer} & All POS & 37\%& 14 & 33\%& 10\\
& VERB & 39\% & 8 &  36\%& 7 \\
& NOUN & 38\% & 16 & 31\%&14\\
& ADJ & 32\%& 11 &36\%& 9\\
& ADV & 33\% & 12 & 38\%& 3\\
\hline
\end{tabular}
\caption{Percentage of the nearest neighbors of hidden states covered by the list of the nearest neighbors of embeddings. }
\label{table:EmbedCoverage}
\end{table*}

\subsection{Syntactic Similarity}
\label{sec:syntax}

Recent extrinsic comparisons of recurrent and non-recurrent architectures on learning syntax \cite{D18_Recurrent,D18-1458_Self-Attention} also motivate a more intrinsic comparison. To this end, we also study the nearest neighbors of hidden states in terms of syntactic similarities. For this purpose, we use the subtree rooting in the smallest phrase constituent above each word, following \newcite{shi-padhi-knight:2016:EMNLP2016}. This way, we will have a corresponding parse tree for each word occurrence in our corpus. We parse our corpus using the Stanford constituent parser \cite{P13-1043}. We POS tag and parse our corpus prior to applying BPE segmentation. Then, after applying BPE, we use the same POS tag and the same subtree of a word for its BPE segments, following \newcite{W16-2209}.

To measure the syntactic similarity between a hidden state and its nearest neighbors, we use PARSEVAL standard metric \cite{evalb} as the similarity metric between the corresponding trees. PARSEVAL computes precision and recall by counting the correct constituents in a parse tree with respect to a gold tree and divide the count with the number of constituent in the candidate parse tree and the gold tree, respectively. 

Figure~\ref{fig:word_of_interest} shows the corresponding word and subtree of a hidden state of interest, and the rest in Figure~\ref{fig:trees} shows the corresponding words and subtrees of its three neighbours. The leaves are substituted with dummy ``*" labels to show that they do not influence the computed tree similarities. We compute the similarity score between the corresponding tree of each word and the corresponding trees of its nearest nighbors. For example, in Figure~\ref{fig:trees} we compute the similarity score between the tree in Figure~\ref{fig:word_of_interest} and each of the other trees.

\subsection{Concentration of Nearest Neighbors}

Each hidden state with its nearest neighbors behaves like a cluster centered around the corresponding word occurrence of the hidden state, whereby the neighboring words give a clearer indication of the captured information in the hidden state. However, this evidence is more clearly observed in some cases rather than others. 

The stronger the similarities that bring the neighbors close to a hidden state, the more focused the neighbors around the hidden state are. Bearing this in mind, we choose to study the relation between the concentration of the neighbors and the information encoded in the hidden states.  

To make it simple but effective, we estimate the population variance of the neighbors' distance from a hidden state as the measure of the concentration of its neighbors. More formally, this is computed as follows:
\begin{equation}
\label{eq:variance}
v_{w_{i,j}} = \frac{1}{n}\sum_{k=1}^{n}(1-x_{k, w_{i,j}})^2
\end{equation} 
Here $n$ is the number of neighbors and $x_{k, w_{i,j}}$ is the cosine similarity score of the $k$th neighbor of word $w$ occurring as the $j$th token of the $i$th sentence.







\section{Empirical Analyses}

We train our systems for English-German and German-English and use our trained model to translate a held-out data of 100K sentences. During translation, we log the hidden state representations together with the corresponding source tokens, their sentence and token indices.

We use the logged hidden states to compute the nearest neighbors of the tokens with frequency of 10 to 2000 in our held-out data. We compute cosine similarity to find the nearest neighbors.

In addition to hidden states, we also log the word embeddings from the same systems and the same trained model. Similar to hidden states, we also use embedding representations to compute the nearest neighbors of words. We have to note that in the case of embedding representations we have one nearest neighbor list for each word whereas for hidden states there is one list for each occurrence of a word.

\setcounter{footnote}{0}

\begin{table*}[thb]
\centering
\small
\begin{tabular}{cccccc}
\hline
Model & POS & English-German& $\sigma^2$ & German-English & $\sigma^2$\\
\hline
\multirow{5}{*}{Recurrent} & All POS & 24\%& 6 & 51\%& 12\\
&VERB & 49\% & 9 & 48\% & 10\\
&NOUN & 19\% & 3 & 28\% & 8\\
& ADJ & 15\% & 2 & 60\% & 12 \\
& ADV & 24\% & 4 & 23\% & 1 \\
\hline
\multirow{5}{*}{Transformer} & All POS & 67\%& 16 & 74\%& 10\\
& VERB & 77\% & 9 & 70\% & 9 \\
& NOUN & 65\% & 18 & 63\% & 13\\
& ADJ & 66\%& 14 & 81\% & 9 \\
& ADV & 74\% & 10 & 35\%& 5\\
\hline
\end{tabular}
\caption{Percentage of the nearest neighbors of hidden states covered by the list of the directly related words to the corresponding word of the hidden states in WordNet. }
\label{table:wordnetCoverage}
\end{table*}

\subsection{Embedding Nearest Neighbors Coverage}

As a first experiment, we measure how many of the nearest neighbors based on the embedding representation would still remain the nearest neighbor of the corresponding hidden state, as described in Section~\ref{sec:hiddenStatesVsEmbeddings}, above.

Table~\ref{table:EmbedCoverage} shows statistics of the coverage by the nearest neighbors based on embeddings in general and based on selected source POS tags for each of our models. To carry out an analysis based on POS tags, we tagged our training data using the Stanford POS tagger \cite{toutanova2003feature}. We convert the POS tags to the universal POS tags and report only for POS tags available in WordNet. We use the same POS tag of a word for its BPE segments, as described in the Section~\ref{sec:syntax}.

The first row of Table~\ref{table:EmbedCoverage} shows that only 18\% and 24\% of the information encoded in the hidden states respectively for English and German is already captured by the word embeddings, in case of our recurrent model. Interestingly, in all cases except ADV, the similarity between the hidden states and the embeddings for the transformer model are much higher, and the increase for nouns is much higher than for the rest. This may be a product of the existence of no recurrence in case of transformer which results in a simpler path from each embedding to the corresponding hidden state. We hypothesize that this means that the recurrent model uses the capacity of its hidden states to encode some other information that is encoded to a lesser extent in the hidden states of the transformer.  

\subsection{WordNet Coverage}

Having observed that a large portion of nearest neighbors of the hidden states are still not covered by the nearest neighbors of the corresponding embeddings, we look for other sources of similarity that causes the neighbors to appear in the list. As the next step, we check to see how many of the neighbors are covered by directly related words of the corresponding word in WordNet.

This does not yield subsets of the nearest neighbors that are fully disjoint with the subset covered by the nearest neighbors from the embedding list. However, it still shows whether this source of similarity is fully covered by the embeddings or whether the hidden states capture information from this source that the embeddings miss.



\begin{figure}[thb]
\includegraphics[scale=0.235]{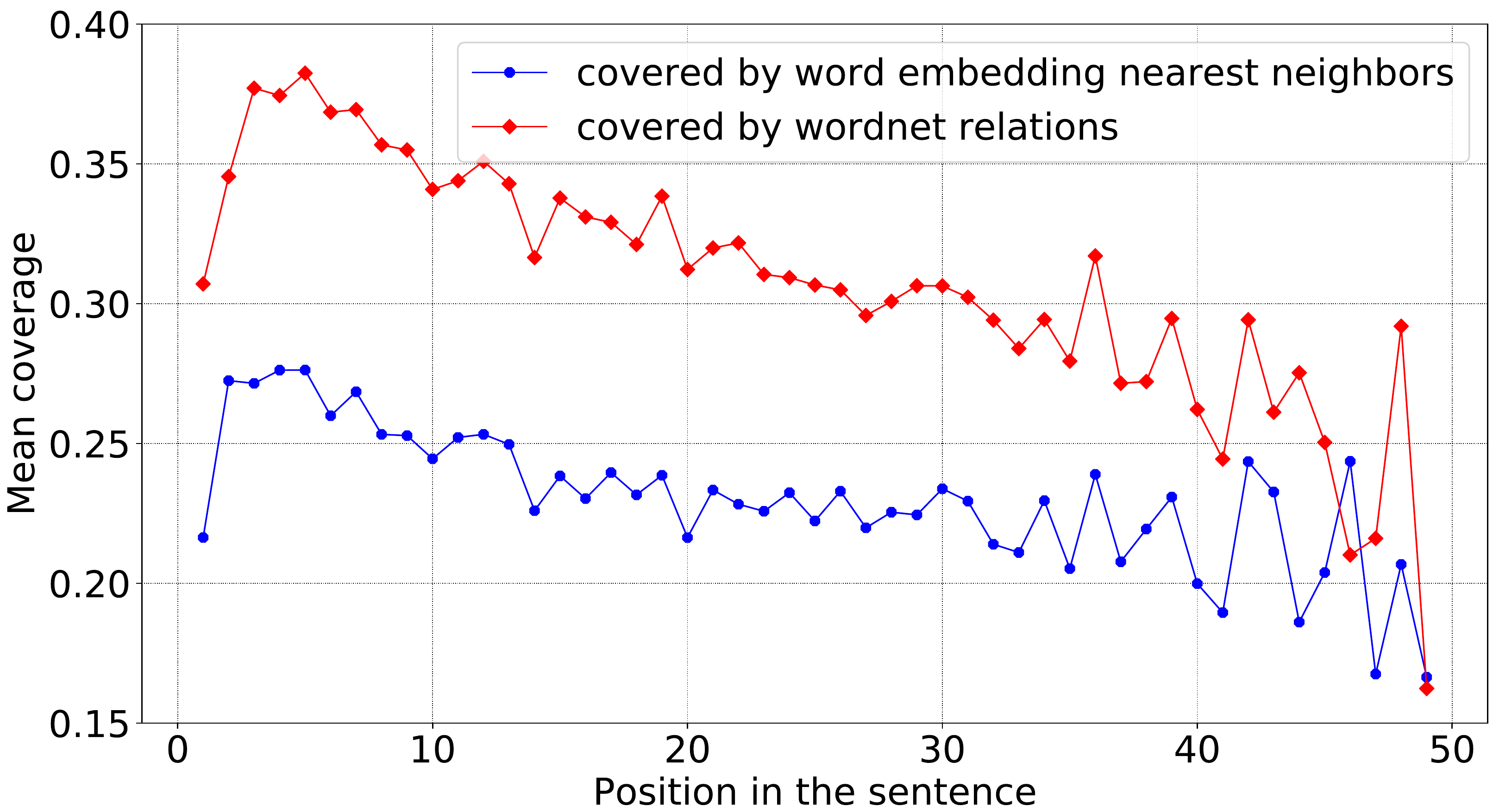}
\caption{The mean coverage per position of the nearest neighbors of hidden states of the recurrent model; (i) by the nearest neighbors of the embedding of the corresponding word (ii) by WordNet related words of the corresponding word of the hidden state.}
\label{fig:wordnetAndEmbeddingCoverage}
\end{figure}

\begin{figure}[thb]
\includegraphics[scale=0.235]{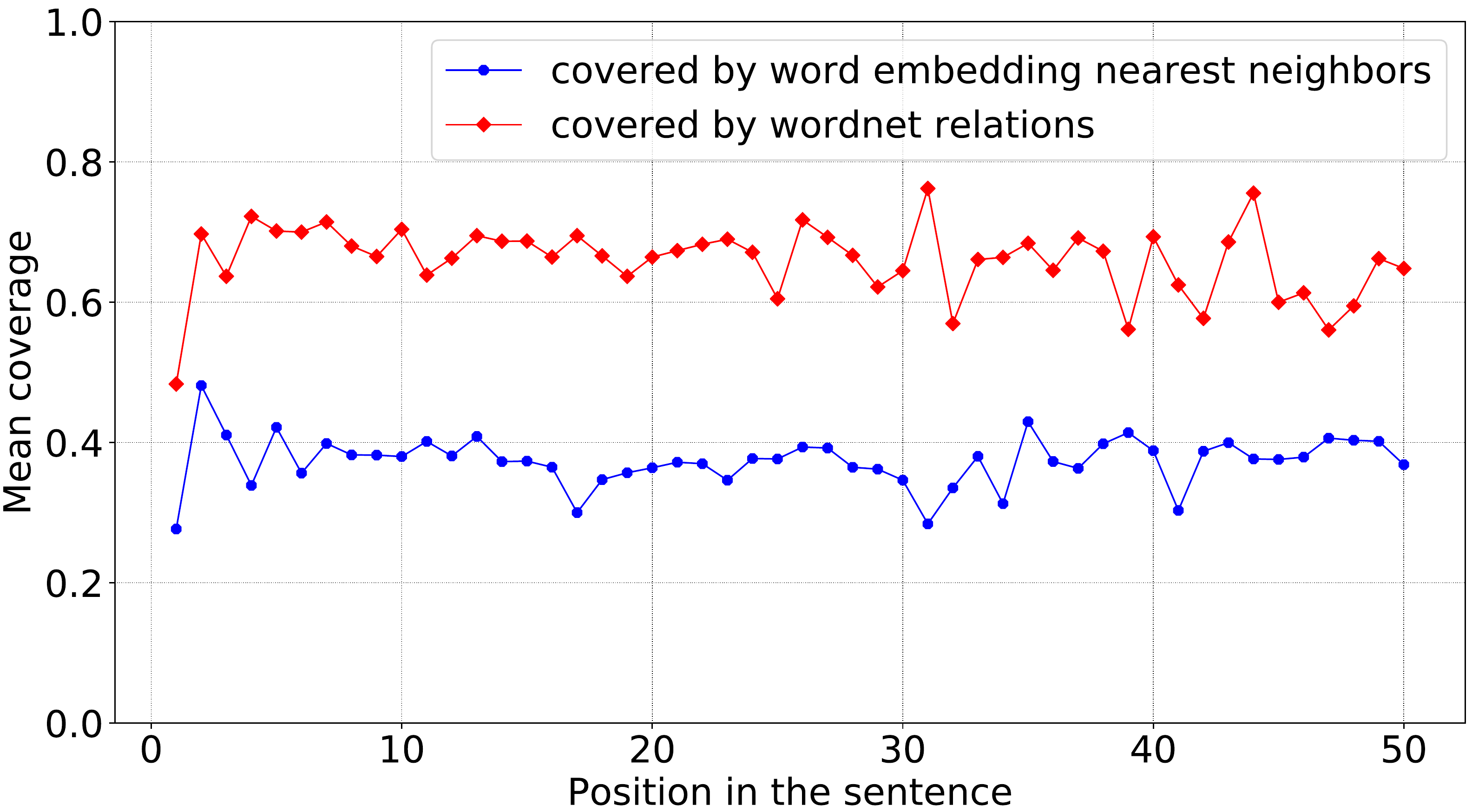}
\caption{The mean coverage per position of the nearest neighbors of hidden states of the transformer model; (i) by the nearest neighbors of the embedding of the corresponding word (ii) by WordNet related words of the corresponding word of the hidden state.}
\label{fig:wordnetAndEmbeddingCoverage_transformer}
\end{figure}

\begin{figure*}[thb]
\centering
\subfloat[English-German, recurrent model \label{fig:variance_recur_1}]{%
\includegraphics[scale=0.235]{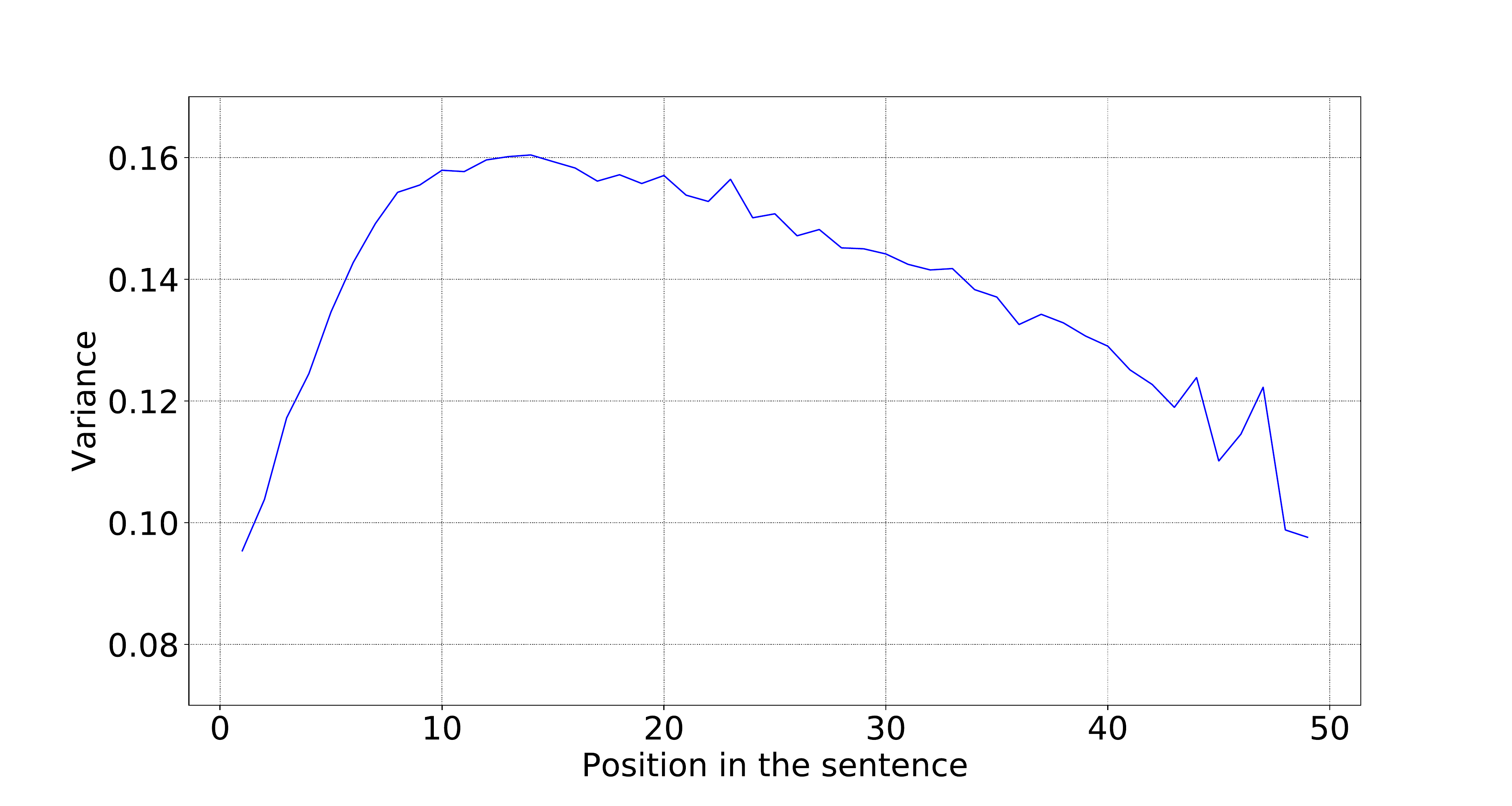}
}
\subfloat[German-English, recurrent model \label{fig:variance_recur_2}]{%
\includegraphics[scale=0.235]{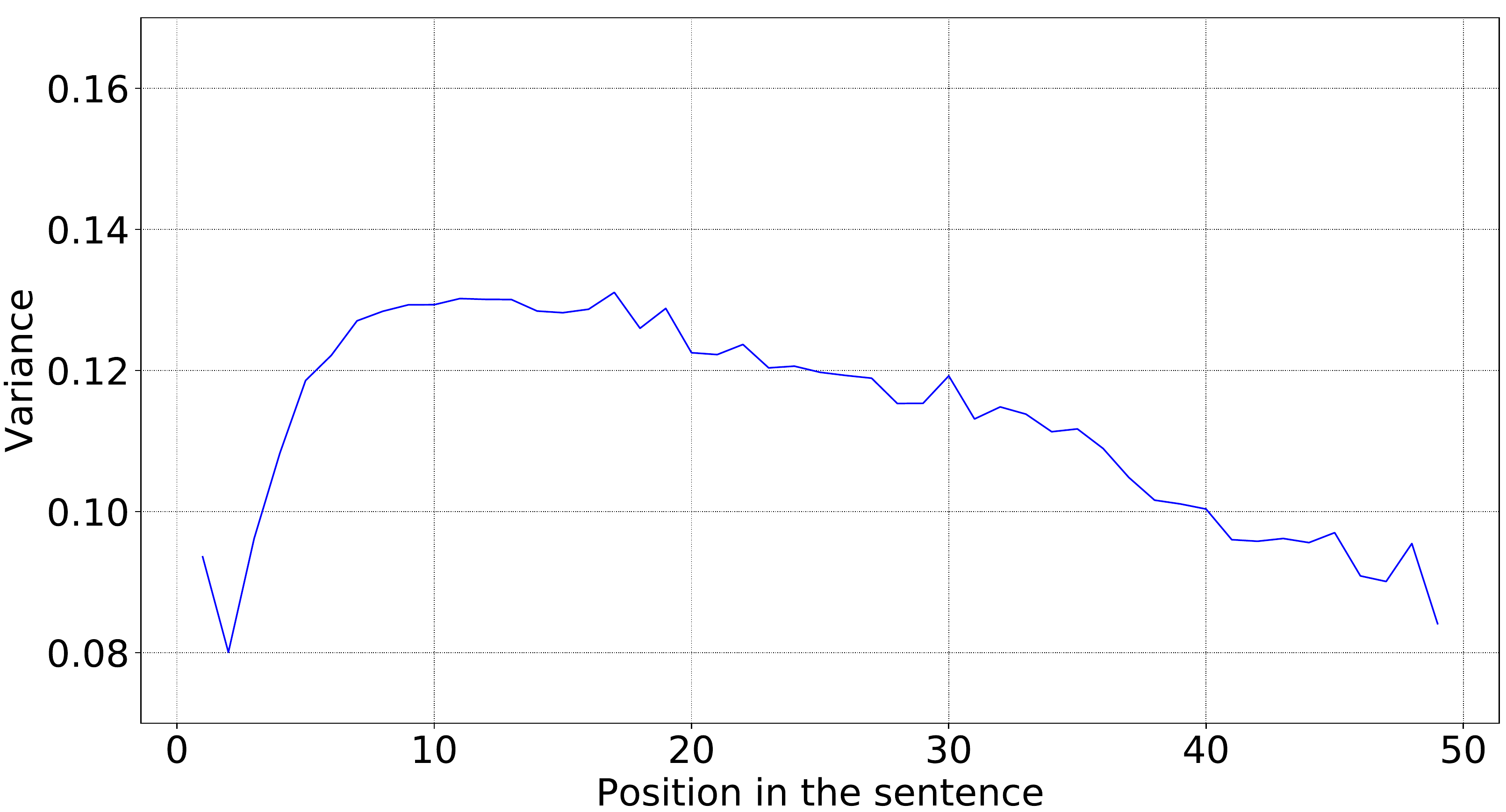}
}

\subfloat[English-German, transformer model \label{fig:variance_transfor}]{%
\includegraphics[scale=0.235]{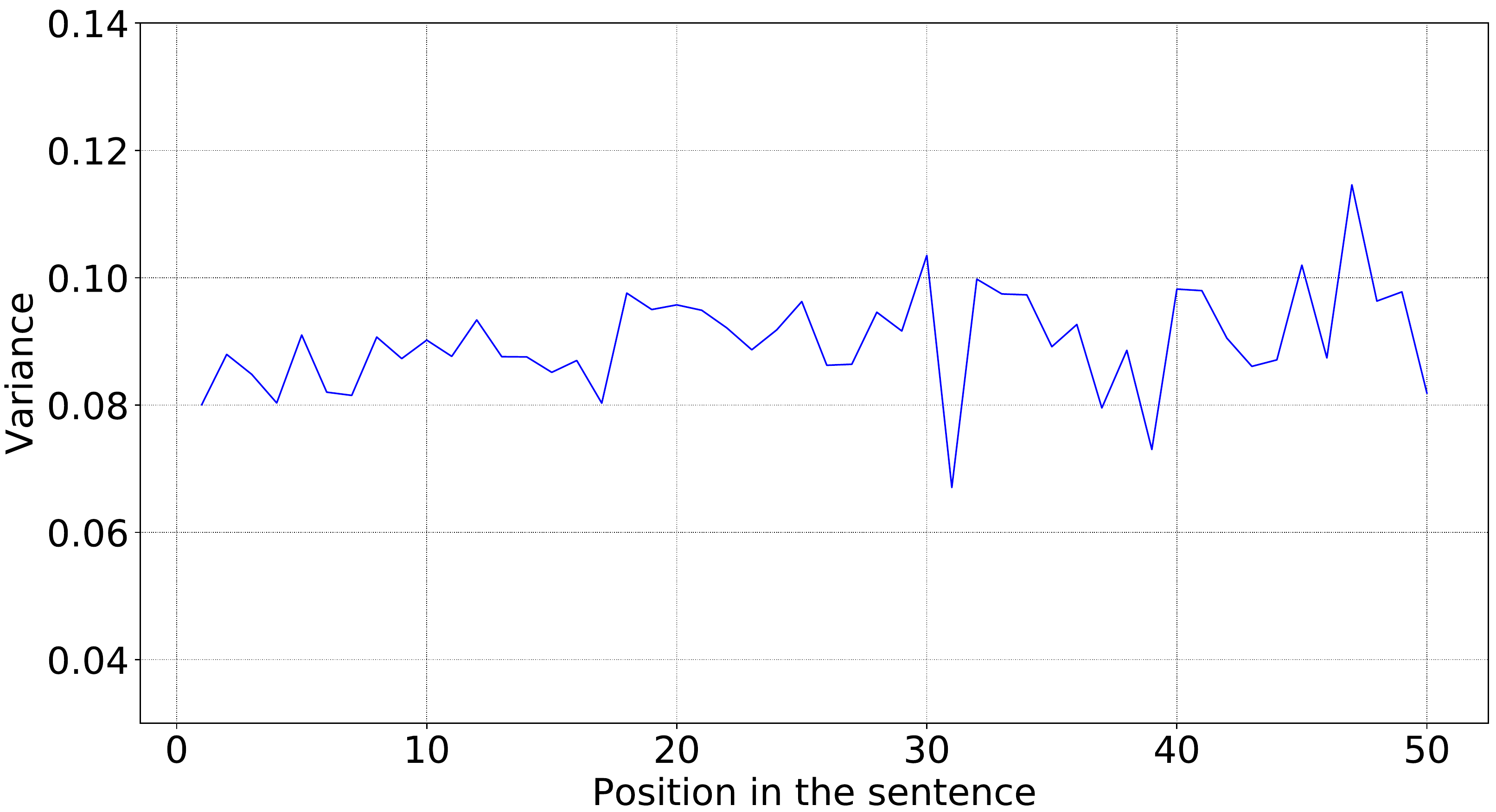}
}
\subfloat[German-English, transformer model \label{fig:variance_transfor_de_en}]{%
\includegraphics[scale=0.235]{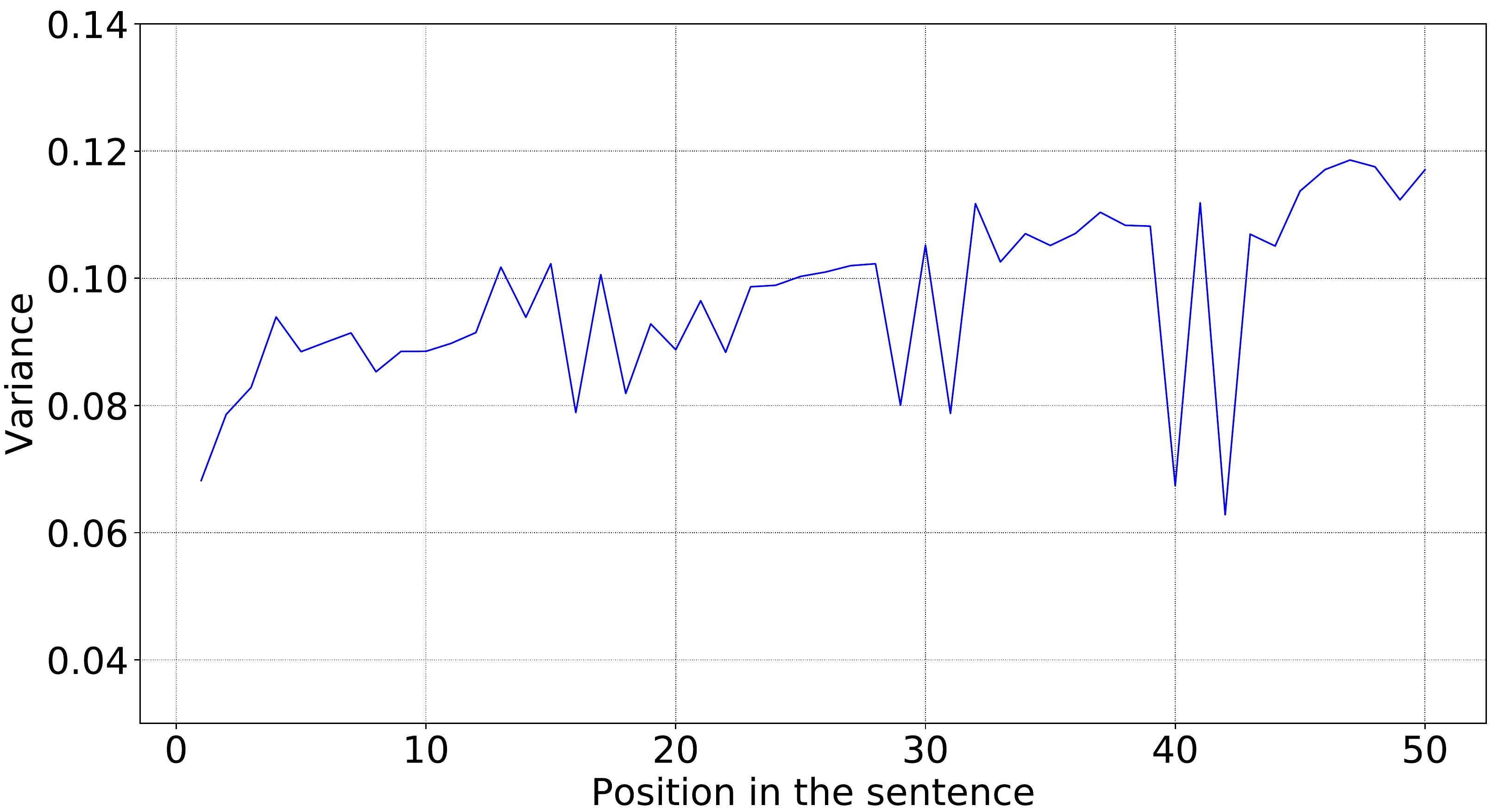}
}
\caption{The average variance of cosine distance scores of the nearest neighbors of words per positions.}
\label{fig:varianceVsPosition}
\end{figure*}

Table~\ref{table:wordnetCoverage} shows the general and the POS-based coverage for our English-German and German-English systems. 
The transformer model again has the lead by a large margin. The jump for nouns is again the highest, as can be seen in Table~\ref{table:EmbedCoverage}. This basically means that more words from the WordNet relations of the word of interest are present in the hidden state nearest neighbors of the word. A simple reason for this could be that the hidden states from transformer capture more word semantic than hidden states of the recurrent model. Or in other words, the hidden states from the recurrent model capture some additional information that brings different words than WordNet relations of the word of interest to its neighborhood.  

To investigate whether recurrency has any direct effect on this, we compute the mean coverage by direct relations in WordNet per position. Similarly, we also compute the mean coverage by embedding neighbors per position. More formally, we write:

\begin{equation}
acp_{j}^{H,W} = \frac{\sum_{i=1}^{m}(cp^{H,W}_{w_{i,j}})}{\left\vert{S_{l(s) \geq j}}\right\vert}
\end{equation}
and 
\begin{equation}
acp_{j}^{H,E} = \frac{\sum_{i=1}^{m}(cp^{H,E}_{w_{i,j}})}{\left\vert{S_{l(s) \geq j}}\right\vert}
\end{equation}
respectively for the mean coverage by WordNet direct relations per position and the mean coverage by embedding neighbors per position.

Here $cp^{H,W}_{w_{i,j}}$ and $cp^{H,E}_{w_{i,j}}$ are the values computed in Equation~\ref{eq:wordnetCov} and~\ref{eq:embeddingCov}, respectively. The function $l(s)$ returns the length of sentence $s$ and $S_{l(s) \geq j}$ is the set of sentences that are longer than or equal to $j$.

\begin{table*}[thb]
\centering
\small
\begin{tabular}{ccccccc}
\multicolumn{6}{c}{English-German}\\
\hline
Model & Precision & Recall & Matched Brackets & Cross Brackets & Tag Accuracy\\
\hline
Recurrent & 0.38 & 0.38 & 0.42 & 0.31 & 0.46 \\
Transformer  & 0.31 & 0.31 & 0.35 & 0.28 & 0.40 \\
\hline
\\
\multicolumn{6}{c}{German-English}\\
\hline
Model & Precision & Recall & Matched Brackets & Cross Brackets & Tag Accuracy\\
\hline
Recurrent &  0.12 & 0.12 & 0.30 & 0.80 & 0.32 \\
Transformer  & 0.11 & 0.11 & 0.28 & 0.77 & 0.31 \\
\hline
\end{tabular}
\caption{Average parse tree similarity (PARSEVAL scores) between word occurrences and their nearest neighbors. Note that the apparent identity of precision and recall values is due to rounding and the very close number of constituents in the corresponding parse tree of words (gold parse trees) and the corresponding parse trees of their nearest neighbors (candidate parse trees).}
\label{tab:syntactic_similarity}
\end{table*}

Figure~\ref{fig:wordnetAndEmbeddingCoverage} shows that the mean coverage in both embedding and WordNet cases is first increasing by getting farther from the left border of a sentence, but it starts to decrease from position 5 onwards, in case of the recurrent model. This is surprising to see the drop in the coverages, taking into account that the model is a bidirectional recurrent model. However, this may be a reflection of the fact that the longer a sentence, the less the hidden states from the recurrent model are encoding information about the corresponding word.

Figure~\ref{fig:wordnetAndEmbeddingCoverage_transformer} shows the same mean coverage for the case of hidden states from the transformer model. No decrease in the coverage per position in the case of transformer, confirms our hypothesis that the lower coverage in case of recurrent models is indeed directly in relation with the recurrency. 

In order to refine the analysis of the positional behaviour of the hidden states, we compute the average variance per position of the cosine distance scores of the nearest neighbors based on hidden states. To compute this value we use the following definition:
\begin{equation}
Av_{j} = \frac{\sum_{i=1}^{m}v_{w_{i,j}}}{\left\vert{S_{l(s) \geq j}}\right\vert}
\end{equation}
Here $v_{w_{i,j}}$ is the variance estimate as defined in Equation~\ref{eq:variance}, $l(s)$ is the function returning the length of sentence $s$ and $S_{l(s) \geq j}$ is the set of sentences that are longer than or equal to $j$ as mentioned earlier. 





Figures~\ref{fig:variance_recur_1} and \ref{fig:variance_recur_2} show the average variance per position for the recurrent model. One can see that the average variance close to the borders is lower than the variance in the middle. This means that the nearest neighbors of the words close to the borders of sentences are more concentrated in terms of similarity score in general. This could mean the information that captures the meaning of those words plays less of a role compared to other information encoded in the corresponding hidden states, especially if we take the information of coverage per position into account.  Interestingly, this does not happen in the case of the transformer model (See Figures~\ref{fig:variance_transfor} and \ref{fig:variance_transfor_de_en} ).

\subsection{Syntactic Similarity}

The difference in the patterns observed so far between recurrent and transformer models (with the coverage of embeddings and the WordNet relations), along with the reported superiority of the recurrent model in capturing structure in extrinsic tasks \cite{D18_Recurrent}, lead us to investigate the syntactic similarity between the words of interest and their nearest neighbors. To this end, we use the approach introduced in the Section~\ref{sec:syntax} to study syntactic similarity.

Table~\ref{tab:syntactic_similarity} shows the average similarity between corresponding constituent subtrees of hidden states and corresponding subtrees of their nearest neighbors, computed using PARSEVAL \cite{evalb}. Interestingly, the recurrent model takes the lead in the average syntactic similarity. This confirms our hypothesis that the recurrent model dedicates more of the capacity of its hidden states, compared to transformer, to capture syntactic structures. It is also in agreement with the results reported on learning syntactic structures using extrinsic tasks \cite{D18_Recurrent}.  We should add that our approach may not fully explain the degree to which syntax in general is captured by each model, but only to the extent to which this is measurable by comparing syntactic structures using PARSEVAL.




\begin{table}[thb]
\centering
\small
\begin{tabular}{ccccc}
\hline
 & \multicolumn{2}{c}{Embedding} & \multicolumn{2}{c}{WordNet}\\
POS & Forward & Backward & Forward & Backward\\
\hline
All & 12\% & 24\% & 18\% & 29\% \\
VERB & 19\% & 36\% & 38\% & 52\%\\
NOUN & 9\% & 21\% & 14\% & 25\%\\
ADJ & 13\% & 22\%& 12\% & 17\%\\
ADV & 28\% & 34\%& 20\%& 23\%\\
\hline
\end{tabular}
\caption{Percentage of the nearest neighbors of hidden states, from the forward and backward layers, that are covered by the list of the nearest neighbors of embeddings and the list of the directly related words in WordNet.}
\label{table:EmbedCoverageLayers}
\end{table}

\begin{figure}[thb]
\centering
\subfloat[Covered by the nearest neighbors of the embedding of the corresponding word of the hidden state.\label{fig:coverage_a}]{%
\includegraphics[scale=0.235]{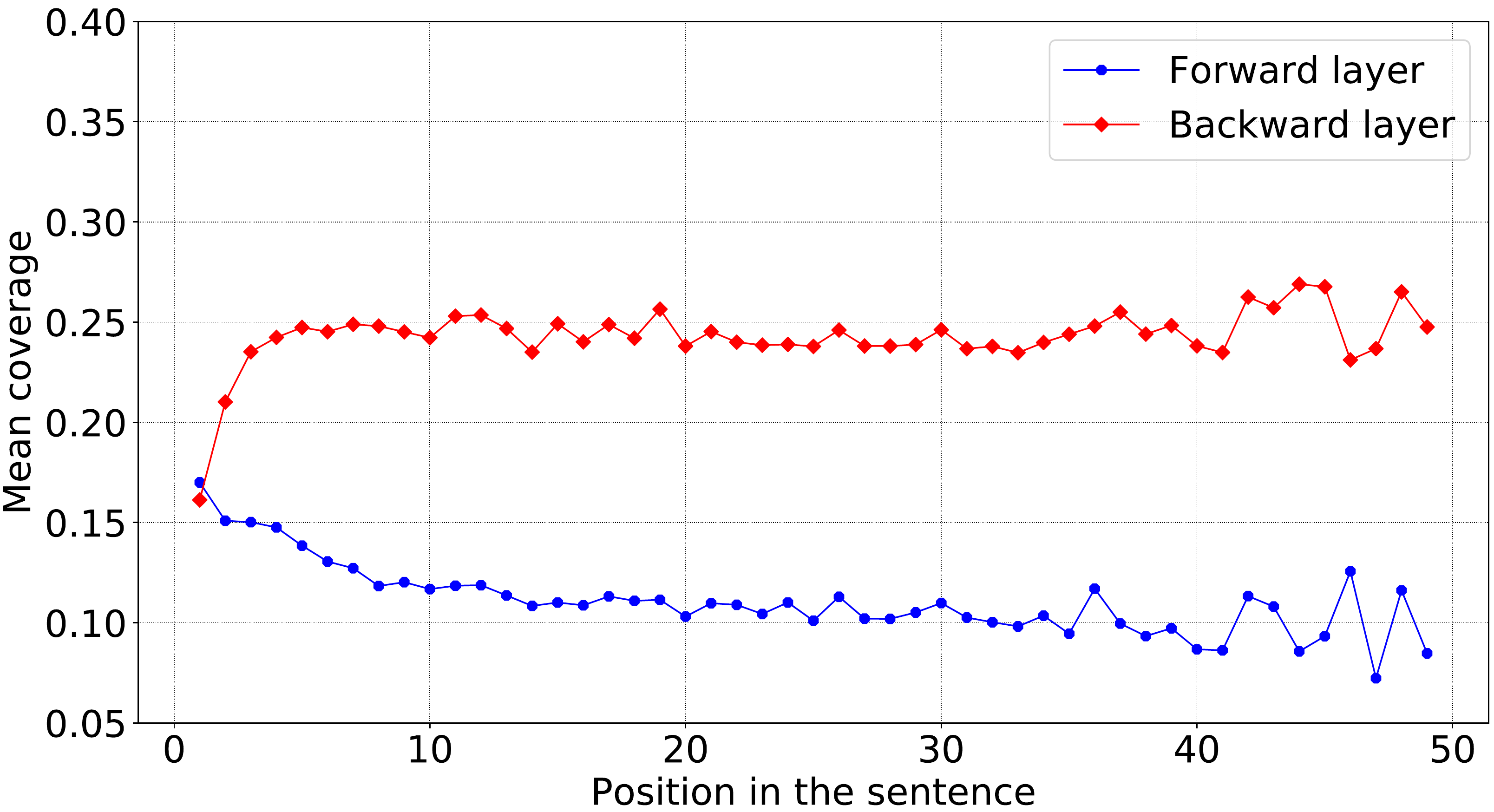}
}

\subfloat[Covered by the directly related words of the corresponding word of the hidden state in WordNet. \label{fig:coverage_b}]{%
\includegraphics[scale=0.235]{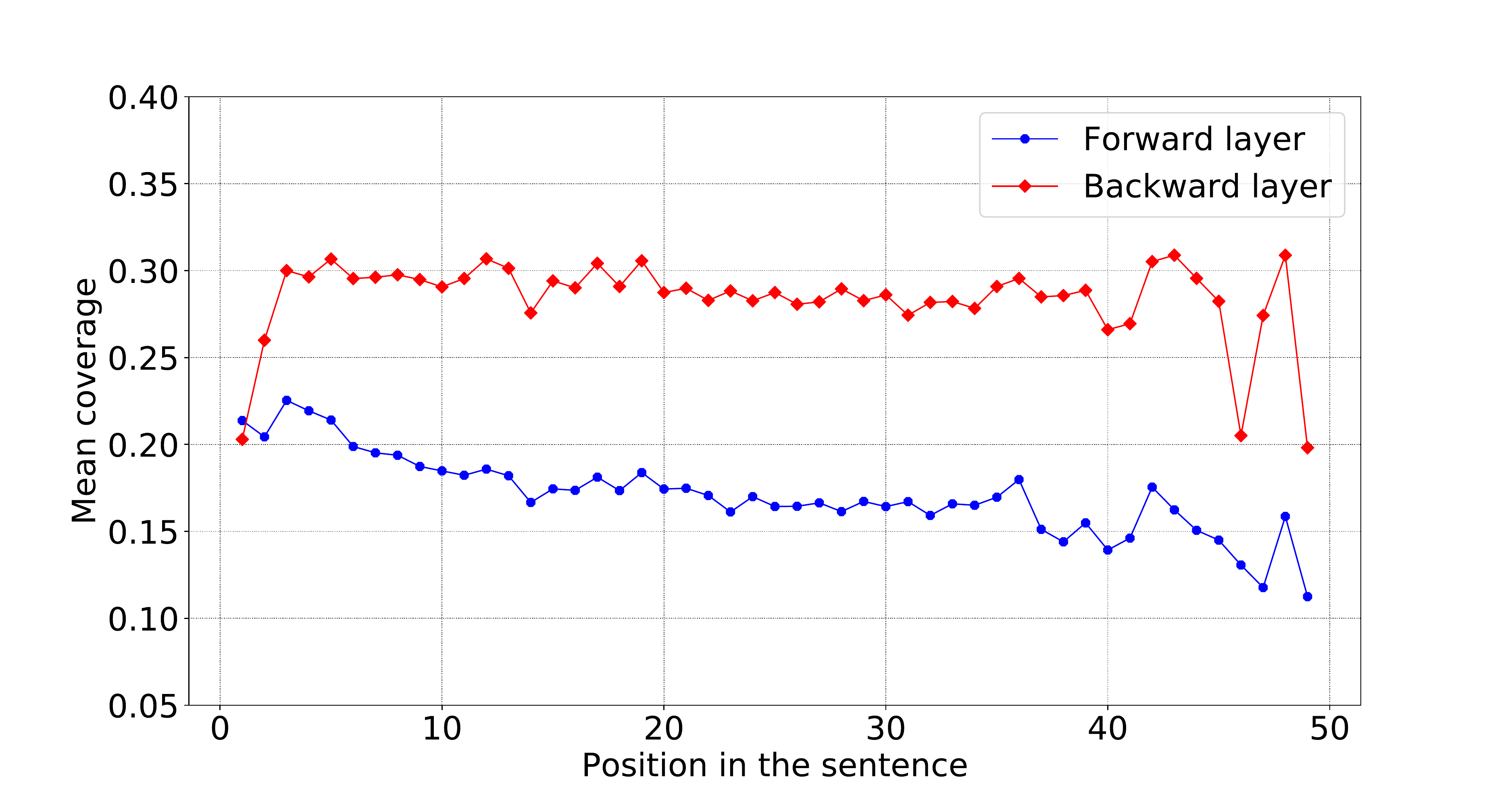}
}
\caption{The mean coverage per position of the nearest neighbors of hidden states from the forward and backward recurrent layers.}
\label{fig:positionCoverage_en_layer12}
\end{figure}





\subsection{Direction-Wise Analyses}

To gain a better understanding of the behaviour of the hidden states in the recurrent model, we repeat our experiments with hidden states from different directions. Note that so far the recurrent hidden states in our experiments were the concatenation of the hidden states from both directions of our encoder.

Table~\ref{table:EmbedCoverageLayers} shows the statistics of embedding coverage and WordNet coverage from the forward and the backward layers. As shown, the coverage of the nearest neighbors of the hidden states from the backward recurrent layer is higher than the nearest neighbors based on those from the forward layer. 

Furthermore, Figure~\ref{fig:positionCoverage_en_layer12} shows the mean coverage per position of the nearest neighbors of hidden states from the forward and the backward recurrent layers. Figure~\ref{fig:coverage_a} shows the mean coverage by the nearest neighbors of the corresponding word embedding of hidden states. As shown for the forward layer the coverage degrades as it goes forward to the end of sentences. However, the coverage for the backward layer, except at the very beginning, almost stays constant through sentences. As shown, the coverage for the backward layer is much higher than the coverage for the forward layer showing that it keeps more information from the embeddings compared to the forward layer. The decrease in the forward layer could mean that it captures more context information when moving forward in sentences and forgets more of the corresponding embeddings.

Figure~\ref{fig:coverage_b} shows the mean coverage by the directly related words, in WordNet, of the corresponding words of hidden states. The difference between the coverage of the nearest neighbors of hidden states from the backward layer comparing to those from the forward layer confirms more strongly that the semantics of words are captured more in the backward layer. This is because here we check the coverage by the directly related words, in the WordNet, of the corresponding words of hidden states.


\section{Conclusion}

In this work, we introduce an intrinsic way of comparing neural machine translation architectures by looking at the nearest neighbors of the encoder hidden states. Using the method, we compare recurrent and transformer models in terms of capturing syntax and lexical semantics. We show that the transformer model is superior in terms of capturing lexical semantics, while the recurrent model better captures syntactic similarity.

We show that the hidden state representations capture quite different information than what is captured by the corresponding embeddings. We also show that the hidden states capture more of WordNet relations of the corresponding word than they capture from the nearest neighbors of the embeddings.

Additionally, we provide a detailed analysis of the behaviour of the hidden states, both direction-wise and for the concatenations. 
We investigate various types of linguistic information captured by the different directions of hidden states in a bidirectional recurrent model.
We show that the reverse recurrent layer captures more lexical and semantic information, whereas the forward recurrent layer captures more long-distance, contextual information. 


\section*{Acknowledgements}

This research was funded in part by the Netherlands Organization for Scientific Research (NWO) under project numbers 639.022.213 and 612.001.218. 

\bibliography{./mtsummit2019}
\bibliographystyle{mtsummit2019}

\end{document}